\documentclass[journal]{IEEEtran}

\usepackage{color,soul}
\sethlcolor{lightgray}

\usepackage[none]{hyphenat} 
\usepackage{array}

\newcolumntype{^}{>{\currentrowstyle}}

\usepackage{booktabs}
\usepackage{multirow}
\usepackage{threeparttable}

\usepackage{graphicx} 
\usepackage{float} 
\usepackage{subfig}
\graphicspath{{figs/}}
\usepackage{tikz}
\usetikzlibrary{shapes.geometric,arrows,matrix,positioning}
\usepackage{pgfplots}

\pgfplotsset{compat=newest} 
\newcounter{plotno}

\usetikzlibrary{pgfplots.groupplots,backgrounds}
\usepgfplotslibrary{units}
\usepackage{tikzscale}

\tikzstyle{io} = [rectangle, minimum width=2cm, minimum height=2cm, text width=2cm, text centered]
\tikzstyle{block} = [rectangle, minimum width=2cm, minimum height=2cm, text width=2cm, text centered, draw=black, line width=0.3mm]
\tikzstyle{arrow_fc} = [line width=0.8mm,->,>=stealth]

\tikzstyle{sum} = [circle, minimum width=0.4cm, draw=black]
\tikzstyle{node1} = [circle, minimum width=0.4cm, draw=black, fill=brown!180]
\tikzstyle{node2} = [circle, minimum width=0.3cm, draw=black, fill=brown!180]
\tikzstyle{node3} = [circle, minimum width=0.2cm, draw=black, fill=brown!180]
\tikzstyle{leaf1} = [rectangle, minimum width=0.2cm, minimum height=0.2cm, draw=black, fill=green]
\tikzstyle{leaf2} = [rectangle, minimum width=0.2cm, minimum height=0.2cm, draw=black, fill=red]
\tikzstyle{arrow} = [line width=0.4mm,-]
\tikzstyle{arrow2} = [line width=0.6mm,->, >=stealth, color=cyan]
\tikzstyle{arrow3} = [line width=0.5mm,->, >=stealth,shorten >= 0.2cm, shorten <= 0.1cm]

\usepackage{mhchem} 
\usepackage{xfrac} 

\usepackage[numbers,sort&compress]{natbib}

\usepackage{setspace}

\usepackage{tikz}
\usepackage[upright]{fourier}
\usepackage{tkz-kiviat,numprint} 

\usetikzlibrary{decorations.pathreplacing, arrows, fit}
\usepackage{graphicx}
\usepackage{url}
\usepackage{booktabs}
\usepackage{textcomp}
\usepackage{epstopdf}
\usepackage{hhline}
\usepackage{xmpmulti}
\usepackage{siunitx}
\usepackage{adjustbox}

\pgfdeclarelayer{background}
\pgfdeclarelayer{foreground}
\pgfsetlayers{background,main,foreground}

\begin{document}
	
\bstctlcite{IEEEexample:BSTcontrol}
%
\title{Semi-supervised Seizure Prediction with Generative Adversarial Networks}
%
%
%

\author{Nhan~Duy~Truong, Levin~Kuhlmann, Mohammad~Reza~Bonyadi, Omid~Kavehei$^{*}$ \thanks{$^{*}$\textit{Corresponding author}: omid.kavehei@sydney.edu.au}
\thanks{N.D.~Truong and O.~Kavehei are with the School of Electrical and Information Engineering and Nano-Neuro-inspired Research Laboratory, The University of Sydney, NSW~2006, Australia.}
\thanks{L.~Kuhlmann is with Centre for Human Psychopharmacology, Swinburne University of Technology, VIC 3122, Australia and Department of Medicine - St. Vincent's and Department of Biomedical Engineering, The University of Melbourne, VIC 3010, Australia}
\thanks{M.R.~Bonyadi is with Centre for Advanced Imaging, University of Queensland, QLD 4072, Australia.}
}%

%


%
%

\markboth{}%
{Shell \MakeLowercase{\textit{et al.}}: Bare Demo of IEEEtran.cls for IEEE Journals}
%



\maketitle

\begin{abstract}
Many outstanding studies have reported promising results in seizure prediction that is considered one of the most challenging predictive data analysis. This is mainly due to the fact that electroencephalogram (EEG) bio-signal intensity is very small, in $\mu$V range, and there are significant sensing difficulties given physiological and non-physiological artifacts. Today the process of accurate epileptic seizure identification and data labeling is done by neurologists. The current unpredictability of epileptic seizure activities together with lack of reliable treatment for patients living with drug resistance forms of epilepsy creates an urgency for research into accurate, sensitive and patient-specific seizure prediction. We believe an advanced, yet computationally efficient, machine learning models, electronic hardware and reliable sensing can be leveraged to enable seizure prediction. In this article, we propose an approach that can make use of not only labeled EEG signals but also the unlabeled ones which is more accessible. We also suggest the use of data fusion to further improve the seizure prediction accuracy. Data fusion in our vision includes EEG signals, cardiogram signals, body temperature and time. We use the short-time Fourier transform on \boldmath$\boldmath28$-s EEG windows as a pre-processing step. A generative adversarial network (GAN) is trained in an unsupervised manner where information of seizure onset is disregarded. The trained Discriminator of the GAN is then used as feature extractor. Features generated by the feature extractor are classified by two fully-connected layers (can be replaced by any classifier) for the labeled EEG signals. This semi-supervised seizure prediction method achieves area under the operating characteristic curve (AUC) of \boldmath$77.68\%$ and \boldmath$75.47\%$ for the CHBMIT scalp EEG dataset and the Freiburg Hospital intracranial EEG dataset, respectively. Unsupervised training without the need of labeling is important because not only it can be performed in real-time during EEG signal recording, but also it does not require feature engineering effort for each patient.
\end{abstract}


%
\IEEEpeerreviewmaketitle

\section{Introduction}
%
%
%
%
\IEEEPARstart{A}{dvances} in deep learning have enabled major improvements in computer vision, language processing and medical applications \cite{krizhevsky2012imagenet,sainath2013cnnspeech,thodoroff2016learning}. In our recent work \cite{Truong2018CNNSZP}, we showed that convolutional neural networks (CNNs) can be used as an effective seizure prediction method. In this work, we exploit deep convolutional generative adversarial network (GAN) \cite{radford2015dcgan} as an unsupervised technique to extract features that can be used for seizure prediction task. The extracted features can be classified by any classifier (neural network with two fully-connected layers in this work). 

Structure of this article is as follows. We first introduce the datasets being used in this work. Next we describe how EEG signals are pre-processed. Then we provide details on GAN and how it can be used as feature extractor for seizure prediction. Lastly, we evaluate our approach and discuss on the results.

\section{Proposed Method}
\subsection{Dataset}
Table~\ref{tbl:szpregan:datasets} summarizes the two datasets being used in this work: CHB-MIT dataset \cite{shoeb2009application} and Freiburg Hospital dataset \cite{EEGFB}. 
CHB-MIT dataset contains scalp EEG (sEEG) data of $23$ pediatric patients with $844$~hours of continuous sEEG recording and $163$ seizures. Scalp EEG signals were captured using $22$ electrodes at sampling rate of $256$~Hz \cite{shoeb2009application}. We define interictal periods that are at least $4$~h away before seizure onset and after seizure ending. In this dataset, there are cases that multiple seizures occur close to each other. For the seizure prediction task, we are interested in predicting the leading seizures. Therefore, for seizures that are less than $30$~min away from the previous one, we consider them as only one seizure and use the onset of leading seizure as the onset of the combined seizure. Besides, we only consider patients with less than $10$ seizures a day for the prediction task because it is not very critical to perform the task for patients having a seizure every $2$~hours on average. With the above definition and consideration, there are $13$ patients with sufficient data (at least $3$ leading seizures and $3$ interictal hours).

The Freiburg dataset consists of intracranial EEG (iEEG) recordings of $21$ patients with intractable epilepsy. Due to lack of availability of the dataset, we are only able to use data from $13$ patients. A sampling rate of $256$~Hz was used to record iEEG signals. In this dataset, there are $6$ recording channels from $6$ selected contacts where three of them are from epileptogenic regions and the other three are from the remote regions. For each patient, there are at least $50$~min preictal data and $24$~h of interictal. More details about Freiburg dataset can be found in \cite{Maiwald2004SPH}.

\begin{table}[htbp]
	\centering
	\caption{Summary of the three datasets used in this paper.\label{tbl:szpregan:datasets}}
	\resizebox{1\columnwidth}{!}{		
		\begin{tabular}{ l*{5}{c}  }			
			\toprule 
			\multirow{2}{3.2em}{\centering Dataset} & \multirow{2}{4em}{\centering EEG type} & \multirow{2}{3.2em}{\centering No. of patients} & \multirow{2}{3.2em}{\centering No. of channels} & \multirow{2}{3.2em}{\centering No. of seizures} & \multirow{2}{3.2em}{\centering Interictal hours} \\ \\					
			\toprule 
			\midrule
			Freiburg & intracranial & $13$ & $6$ & $59$ & $311.4$ \\
			CHB-MIT & scalp & $13$ & $22$ & $64$ & $209$ \\			
			\bottomrule			
		\end{tabular}	
	}
\end{table}

\subsection{Pre-processing}
\label{preprocessing}
\begin{figure}[h]
\subfloat[]{%
	\includegraphics[width=0.9\columnwidth]{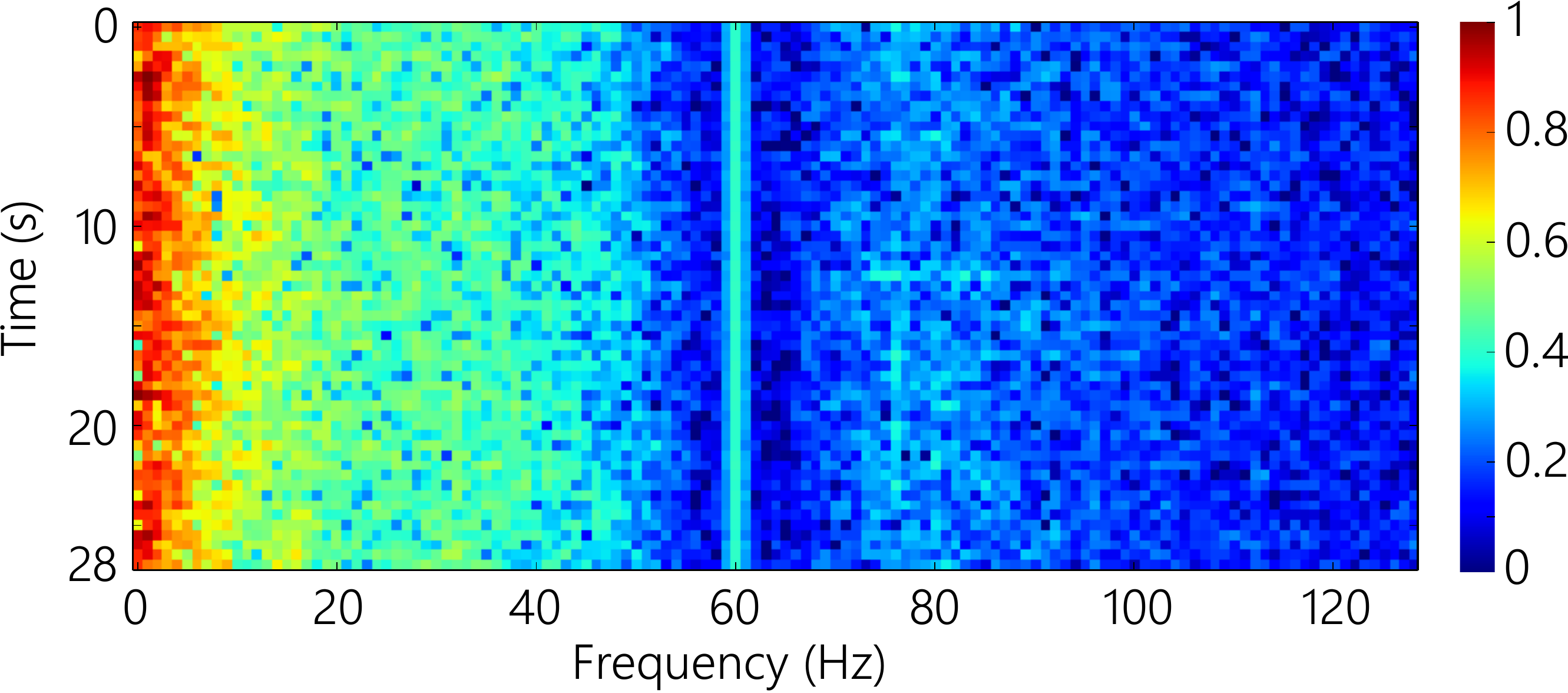}%
}

\subfloat[]{%
	\includegraphics[width=0.8\columnwidth]{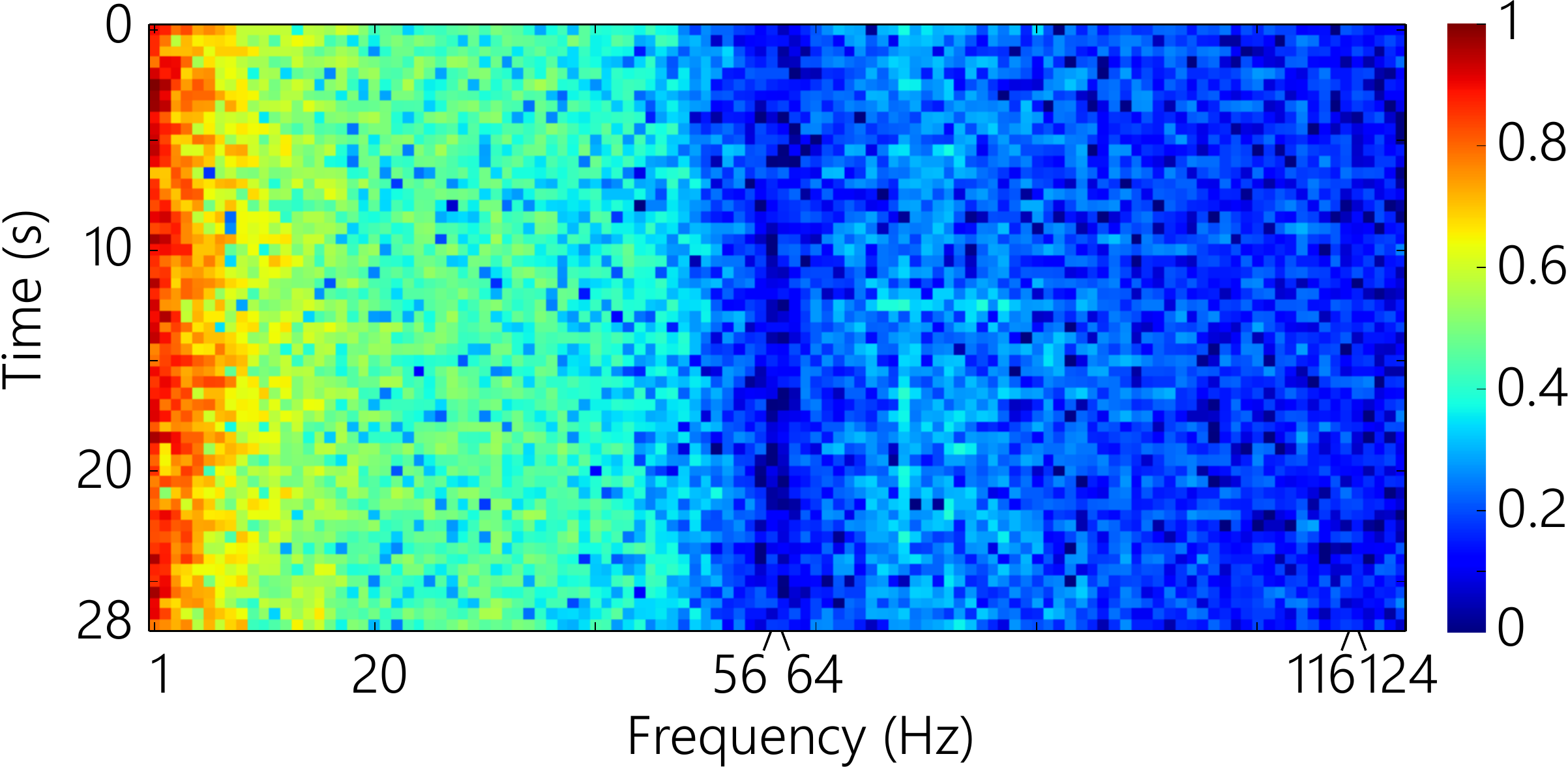}%
}


\caption{(a) Example STFT of a $28$~second window. (b) Same window after removing line noise.}
\label{fig:szpregan:stft}
\end{figure}

Since we will use a Generative Neural Network (GAN) architecture with three de-convolution layers, dimensions of GAN's input must be divisible by $2^{3}$, except the number of channels. Specific to CHBMIT dataset, there are some patients that have less than $22$ channels of recording EEG due to changes in electrodes. Particularly, Pat13 and Pat17 have only $17$ available channels; Pat4, Pat9 have $20$, $21$ channels, respectively. Since we are interested in whether GAN can be effectively trained with non-patient specific data, all patients must have the same number of channels so that data from all patients can be combined. We follow approach in \cite{Truong2017ACS} to select $16$ channels for each patient in CHBMIT dataset. With regards to CHB-MIT and Freiburg datasets, we use Short-Time Fourier Transform (STFT) to translate $28$~seconds of time-series EEG signal into two dimensional matrix comprised of frequency and time axes. For the STFT, we use cosine window of $1$~second length and $50\%$ overlap. Most of EEG recordings were contaminated by power line noise at $60$~Hz (see Fig.~\ref{fig:szpregan:stft}a) for CHB-MIT dataset and $50$~Hz for Freiburg dataset. The power line noise can be removed by excluding components at frequency range of $47$--$53$~Hz and $97$--$103$~Hz if power frequency is $50$~Hz and components at frequency range of $57$--$63$~Hz and $117$--$123$~Hz for power line frequency of $60$~Hz. The DC component (at $0$~Hz) was also removed. Fig.~\ref{fig:szpregan:stft}b shows the STFT of a $28$-s window after removing power line noise. We also trim components at the last two frequencies $127$--$128$~Hz to have the final dimension of each pre-processed $28$~s be $n \times 56 \times 112$, where $n=16$ for CHBMIT dataset and $n=6$ for Freiburg dataset. 

\subsection{Adversarial Neural Network}

\begin{figure*}[h]
\centering
\includegraphics[width=0.75\textwidth]{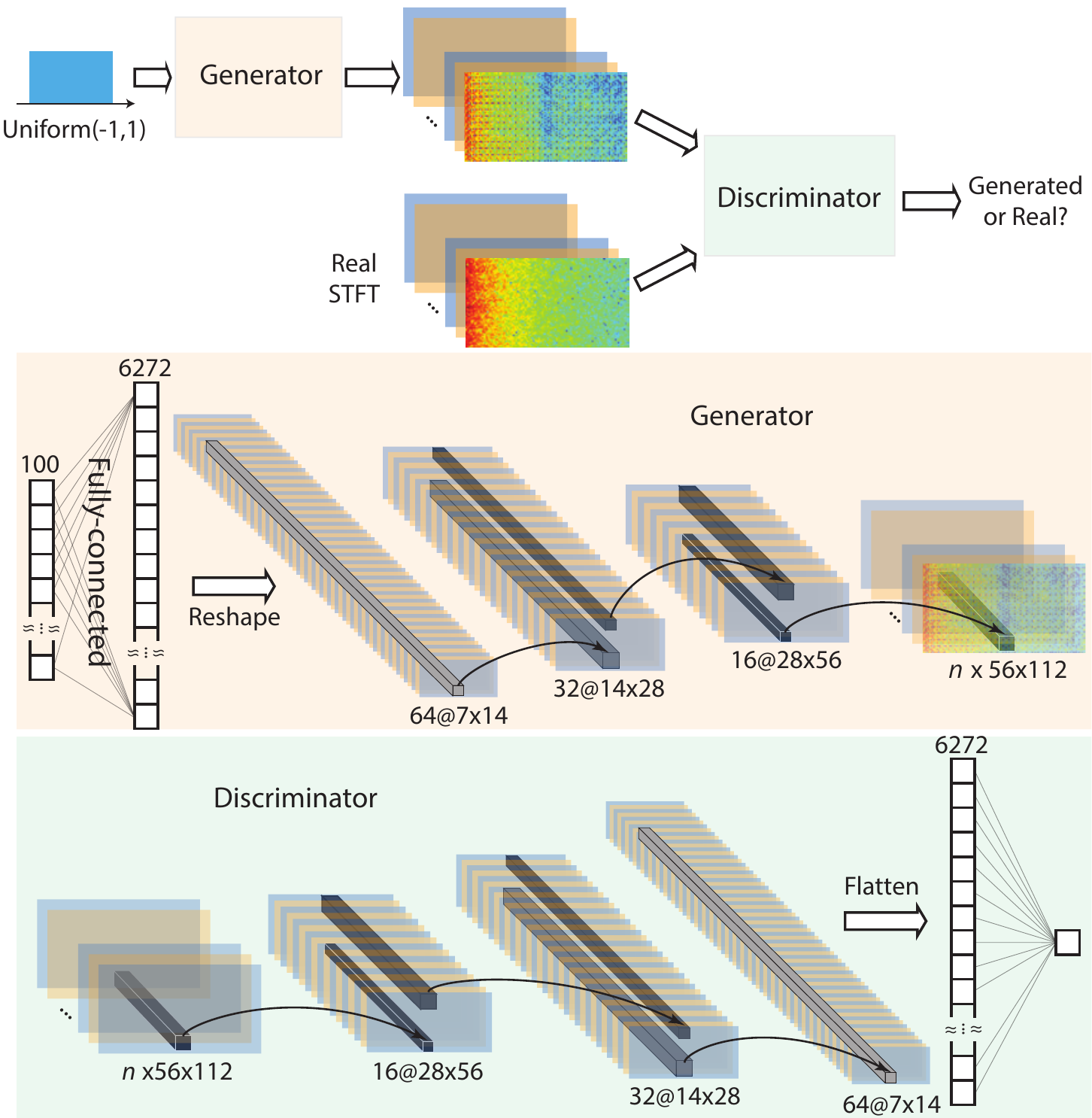}
\caption{The Generator takes a random sample of $100$ data points from a uniform distribution $\mathcal{U}(-1,1)$ as input. The input is fully-connected with a hidden layer with output size of $6272$ which is then reshaped to $64 \times 7 \times 14$. The hidden layer is followed by three de-convolution layers with filter size $5 \times 5$, stride $2 \times 2$. Numbers of filters of the three de-convolution layers are $32$, $16$ and $n$, respectively. The Discriminator consists of three convolution layers with filter size $5 \times 5$, stride $2 \times 2$. Numbers of filters of the three convolution layers are $16$, $32$ and $64$, respectively.}
\label{fig:szpregan:gan}
\end{figure*}

In this paper, we use a Generative Adversarial Network (GAN) \cite{goodfellow2014generative} as depicted in Fig.~\ref{fig:szpregan:gan} as an unsupervised feature extraction technique. Note that here we explain for the CHBMIT dataset. The same explanation is applied for the other two datasets with the change in input dimension as mentioned in Section~\ref{preprocessing}. The Generator takes a random sample of $100$ data points from a uniform distribution $\mathcal{U}(-1,1)$ as input. The input is fully-connected with a hidden layer with output size of $6272$ which is then reshaped to $64 \times 7 \times 14$. The hidden layer is followed by three de-convolution layers with filter size $5 \times 5$, stride $2 \times 2$. Numbers of filters of the three de-convolution layers are $32$, $16$ and $n$, respectively. Outputs of the Generator have the same dimension with STFT of $28$~seconds EEG signals. The Discriminator, on the other hand, is configured to discriminate the generated EEG signals from the original ones. The Discriminator consists of three convolution layers with filter size $5 \times 5$, stride $2 \times 2$. Numbers of filters of the three convolution layers are $16$, $32$ and $64$, respectively. During training, the Generator tries to generate signals that "look" like the original ones while the Discriminator is optimized to detect those generated signals. As a result, the Discriminator learns how to extract unique features in the original EEG signals by adjusting its parameters in the three convolution layers. This training process is unsupervised because we do not provide labels (preictal or interictal) to the network.

\subsection{Convolutional neural network}
After training the GAN, we add two fully-connected layers with sigmoid activation and output sizes of $256$ and $2$, respectively, after the trained convolution layers in GAN's Discriminator to form a Convolutional Neural Network (CNN) for seizure prediction task. The former fully-connected layer uses sigmoid activation function while the latter uses soft-max activation function. Both of the two fully-connected layers have drop-out rate of $0.5$. We then train the CNN as normal except all trained convolution layers are kept unchanged. In this configuration, the three convolution blocks that are ready trained play as feature extractor, the two fully-connected layers play as a classifier. Our model training is performed on a NVIDIA P100 graphic card using Tensorflow 1.4.0 framework. We also apply a practice proposed in (cite-seizure-prediction-paper) to prevent over-fitting during training the CNN. Specifically, we choose $25\%$ later preictal and interictal samples from the training set to monitor if over-fitting occurs and use the rest to train the network. Dataset balancing technique proposed in (cite-seizure-prediction-paper) is also applied in this paper.

\begin{figure*}[h]
\centering
\includegraphics[width=0.98\textwidth]{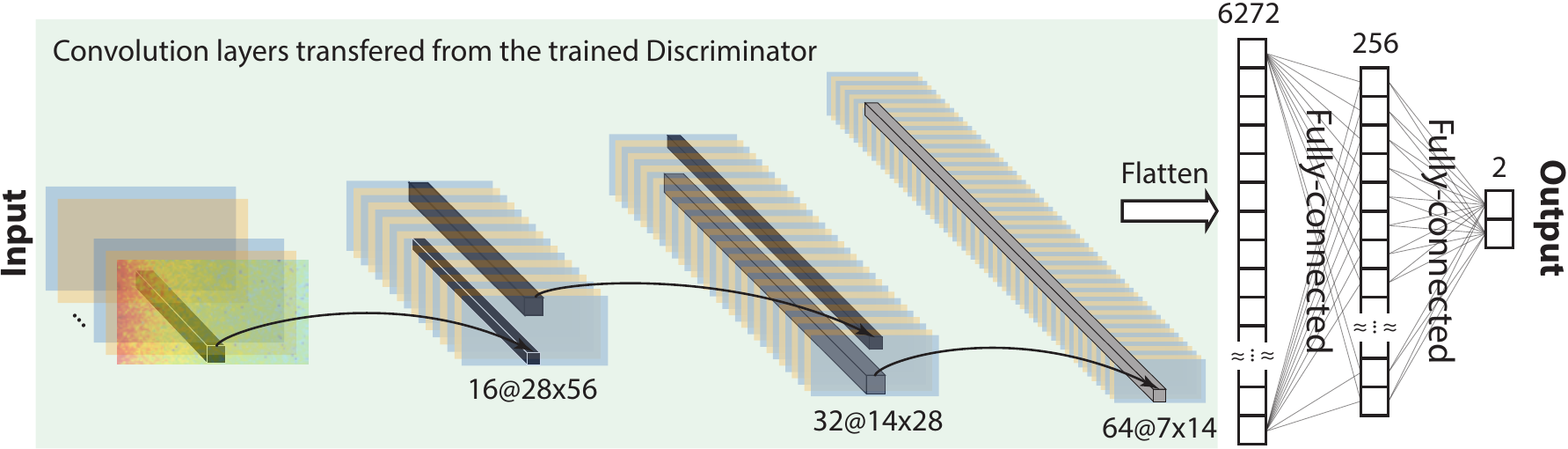}
\caption{Convolutional neural network architecture. This illustration is applied to Freiburg and CHB-MIT datasets.  
Input are STFT transforms of $28$s windows of raw EEG signals. The three convolution blocks are transfered from the trained GAN and are kept unchanged during training the CNN. Features extracted by the three convolution blocks are flatten and connected to $2$ fully-connected layers with output sizes $256$ and $2$, respectively. The former fully-connected layer uses sigmoid activation function while the latter uses soft-max activation function. Both of the two fully-connected layers have drop-out rate of $0.5$. In this configuration, the three convolution blocks that are ready trained play as feature extractor, the two fully-connected layers play as a classifier.}
\label{fig:szpregan:cnn}
\end{figure*}

\subsection{System evaluation}

\begin{figure}[h]
\centering
\includegraphics[width=1\columnwidth]{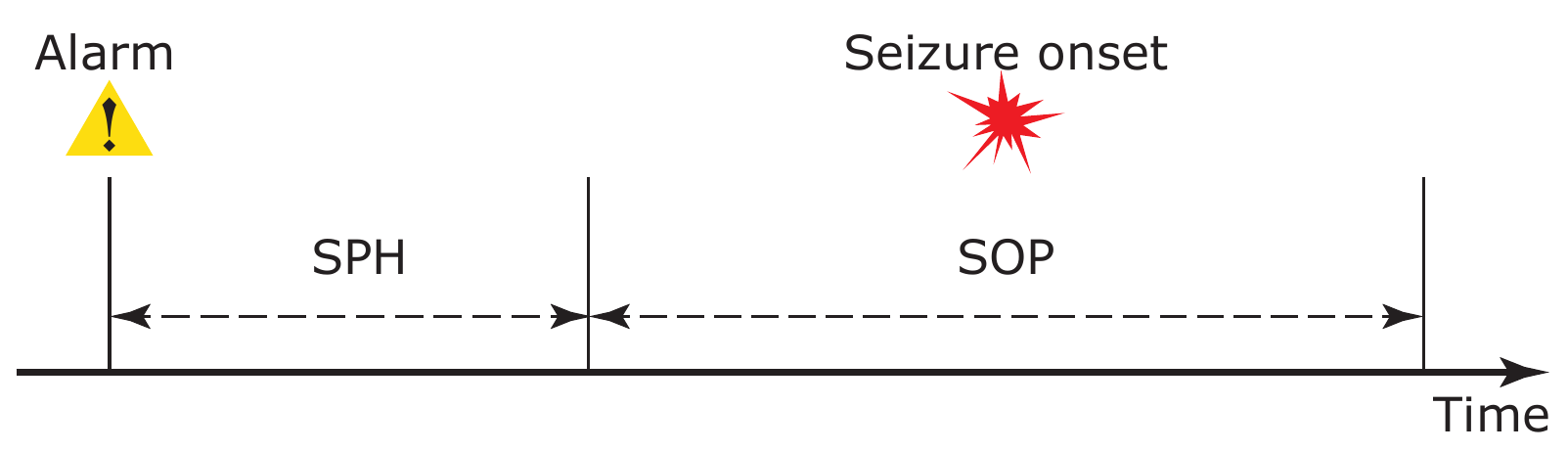}
\caption{Definition of seizure occurrence period (SOP) and seizure prediction horizon (SPH). For a correct prediction, a seizure onset must be is after the SPH and within the SOP.}
\label{fig:szpregan:sph}
\end{figure}

Seizure prediction horizon (SPH) and seizure occurrence period (SOP) need to be defined before estimating the system's performance. In this paper, we follow the definition of SOP and SPH that was proposed in \cite{Maiwald2004SPH} (see~Fig.~\ref{fig:szpregan:sph}). SOP is the interval where the seizure is expected to occur. The time period between the alarm and beginning of SOP is called SPH. For a correct prediction, a seizure onset must be after the SPH and within the SOP. Likewise, a false alarm rises when the prediction system returns a positive but there is no seizure occurring during SOP. When an alarm rises, it will last until the end of the SOP. Regarding clinical use, SPH must be long enough to allow sufficient intervention or precautions (SPH is also called intervention time \cite{BouAssi2017szpred}). In contrast, SOP should be not too long to reduce the patient's anxiety. 

We use area under the receiver operating characteristics curve (AUC) with SPH of $5$~min and SOP of $30$~min. To have a robust evaluation, we follow a leave-one-out cross-validation approach for each subject. If a subject has $N$ seizures, $(N-1)$ seizures will be used for training and the withheld seizure for validation. This round is repeated $N$ times so all seizures will be used for validation exactly one time. Interictal segments are randomly split into $N$ parts. $(N-1)$ parts are used for training and the rest for validation. The $(N-1)$ parts are further split into monitoring and training sets to prevent over-fitting (cite prediction paper here).

\section{Results}
\begin{figure*}[h]
\centering
\includegraphics[width=0.85\textwidth]{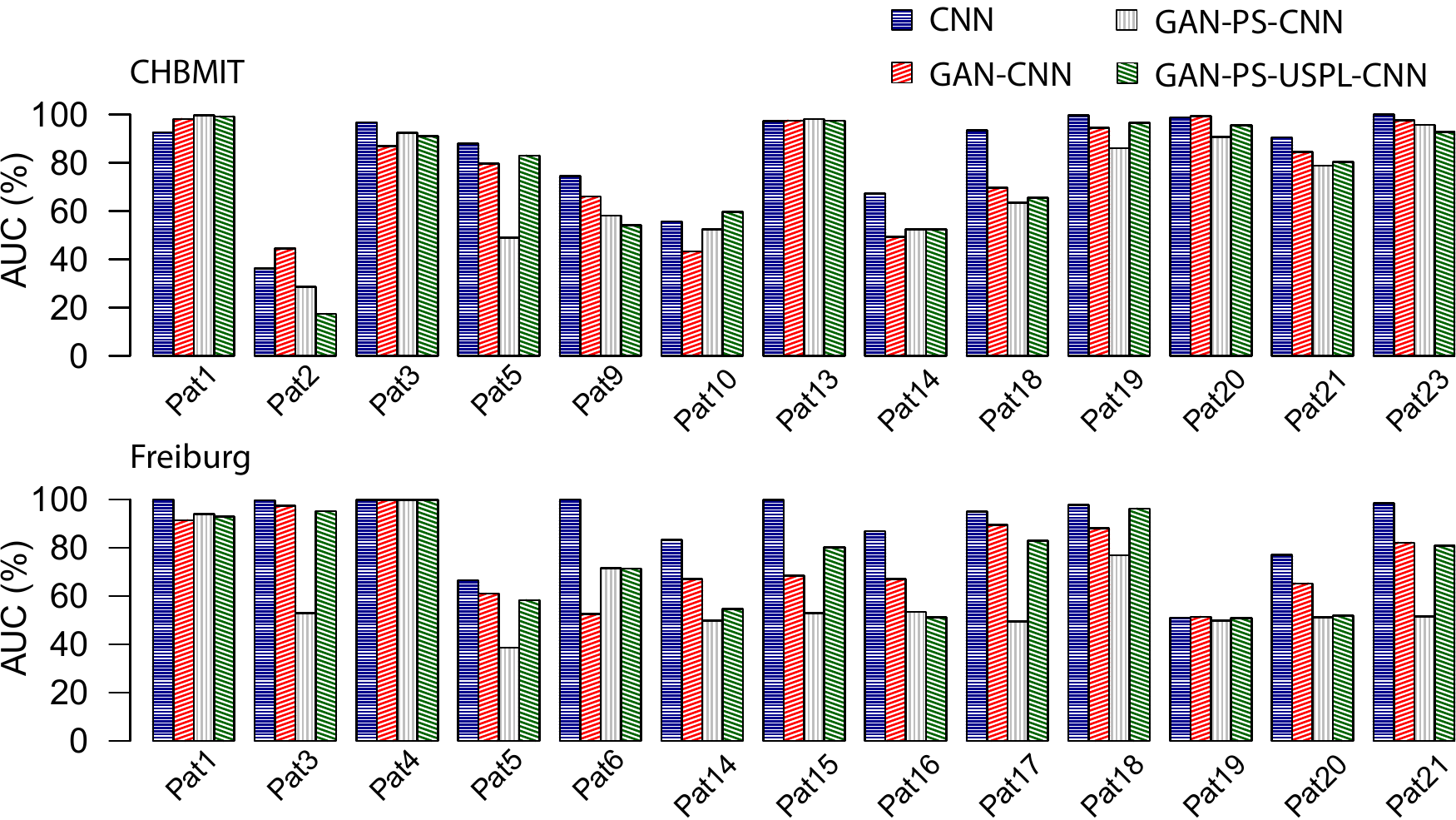}
\caption{Seizure prediction performance for the CHBMIT dataset (upper) and the Freiburg Hospital dataset (lower). Four methods are evaluated: (1) CNN: convolutional neural network, (2) GAN-CNN: unsupervised feature extraction using GAN and classification performed by a CNN, (3) GAN-PS-CNN similar to (2) but GAN is done patient-specific, (4) GAN-PS-OSPL-CNN: similar to (3) but $10 \times$ over-sampling of samples is performed when training GAN. }
\label{fig:szpregan:auc}
\end{figure*}

In this section, we test our approach with two datasets: CHB-MIT sEEG dataset and Freiburg iEEG dataset. 
SOP~=~$30$~min and SPH~=~$5$~min were used in calculating all metrics in this paper. Each fold of leave-one-out cross-validation was executed twice and average results with standard deviations were reported. Fig.~\ref{fig:szpregan:auc} summarizes seizure prediction results with SOP of $30$~min and SPH of $5$~min. We investigate the system performance in three scenarios: (1) GAN is trained with data of all patients combined (from the same dataset), (2) GAN is trained in a patient-specific fashion, and (3) GAN is trained in a patient-specific fashion with improvement. In scenario (3), similar to dataset balancing technique proposed in (cite-seizure-prediction-paper), we generate extra samples from existing ones. As a result, training set in scenario (3) is ten times larger compared to the one in scenario (2). The results are shown in Tables~\ref{tbl:szpregan:results_CHBMIT}-\ref{tbl:szpregan:results_FB} and Fig.~\ref{fig:szpregan:auc}. Compared to the fully supervised CNN, GAN-CNN introduces $\approx 6\%$ and $\approx 12\%$ loss in AUC for the CHBMIT sEEG dataset and the Freiburg Hospital iEEG dataset, respectively. When GAN is trained per patient (GAN-PS-CNN), the average AUC drops further to $72.63\%$ and $60.91\%$ for the two datasets. This can be explained by the limited amount of data from each patient. By applying $10 \times$ over-sampling (GAN-PS-OSPL-CNN), the average AUC is boosted to $75.66\%$ and $74.33\%$ for the CHBMIT dataset and the Freiburg Hospital dataset, respectively, which are $1$--$2\%$ lower than those of GAN-CNN.

\begin{table}[htbp]
	\centering
	\caption{Seizure prediction performance for the CHBMIT dataset.\label{tbl:szpregan:results_CHBMIT}}
	\resizebox{1\columnwidth}{!}{		
		\begin{tabular}{ l*{4}{c}  }			
			\toprule 
			\multirow{2}{2em}{\centering Patient} & \multirow{2}{4em}{\centering CNN} & \multirow{2}{4em}{\centering GAN-CNN} & \multirow{2}{4em}{\centering GAN-PS CNN} & \multirow{2}{4em}{\centering GAN-PS USPL-CNN} \\ \\					
			\toprule 
			\midrule
			Pat1 & $92.48$ & $98.09$ & $99.52$ & $99.13$ \\
            Pat2 & $36.16$ & $44.47$ & $28.52$ & $17.34$ \\
            Pat3 & $96.66$ & $86.79$ & $92.43$ & $90.91$ \\
            Pat5 & $87.8$ & $79.62$ & $48.83$ & $82.9$ \\
            Pat9 & $74.41$ & $65.87$ & $57.99$ & $54$ \\
            Pat10 & $55.59$ & $43.17$ & $52.38$ & $59.63$ \\
            Pat13 & $97.21$ & $97.42$ & $98.04$ & $97.35$ \\
            Pat14 & $67.16$ & $49.22$ & $52.28$ & $52.34$ \\
            Pat18 & $93.29$ & $69.54$ & $63.27$ & $65.44$ \\
            Pat19 & $99.48$ & $94.53$ & $85.93$ & $96.36$ \\
            Pat20 & $98.67$ & $99.21$ & $90.7$ & $95.43$ \\
            Pat21 & $90.47$ & $84.38$ & $78.71$ & $80.17$ \\
            Pat23 & $99.9$ & $97.55$ & $95.59$ & $92.6$ \\
            \midrule
            Average & $83.79$ & $77.68$ & $72.63$ & $75.66$ \\	
            \bottomrule			
		\end{tabular}	
	}
\end{table}

\begin{table}[htbp]
	\centering
	\caption{Seizure prediction performance for the Freiburg Hospital dataset.\label{tbl:szpregan:results_FB}}
	\resizebox{1\columnwidth}{!}{		
		\begin{tabular}{ l*{4}{c}  }			
			\toprule 
			\multirow{2}{2em}{\centering Patient} & \multirow{2}{4em}{\centering CNN} & \multirow{2}{4em}{\centering GAN-CNN} & \multirow{2}{4em}{\centering GAN-PS CNN} & \multirow{2}{4em}{\centering GAN-PS USPL-CNN} \\ \\					
			\toprule 
			\midrule
			Pat1 & $100$ & $91.43$ & $94.02$ & $92.78$ \\
            Pat3 & $99.59$ & $97.44$ & $52.89$ & $95.13$ \\
            Pat4 & $99.93$ & $99.92$ & $99.88$ & $99.88$ \\
            Pat5 & $66.58$ & $61.04$ & $38.6$ & $58.28$ \\
            Pat6 & $100$ & $52.58$ & $71.51$ & $71.27$ \\
            Pat14 & $83.28$ & $67.01$ & $49.86$ & $54.6$ \\
            Pat15 & $99.95$ & $68.5$ & $52.88$ & $80.18$ \\
            Pat16 & $86.81$ & $67.01$ & $53.44$ & $51.17$ \\
            Pat17 & $94.92$ & $89.44$ & $49.49$ & $82.91$ \\
            Pat18 & $97.69$ & $87.99$ & $76.9$ & $96.25$ \\
            Pat19 & $50.97$ & $51.35$ & $49.77$ & $50.93$ \\
            Pat20 & $77.02$ & $65.24$ & $51.11$ & $51.91$ \\
            Pat21 & $98.4$ & $82.14$ & $51.51$ & $80.94$ \\
            \midrule
            Average & $88.86$ & $75.47$ & $60.91$ & $74.33$ \\		
            \bottomrule			
		\end{tabular}	
	}
\end{table}

\section{Discussion}
We have shown that feature extraction for seizure prediction can be done in an unsupervised way. Though the overall AUC degraded by $\approx 6\%$ for CHBMIT dataset and $\approx 12\%$ for Freiburg Hospital dataset, our unsupervised feature extraction can help to minimize the EEG labeling task that is costly and time consuming. Specifically, unlabeled EEG signals are used to train the GAN. The trained GAN plays as a feature extractor. Extracted features from labeled EEG data (that can be much smaller than unlabeled one) can be fed to any classifier (two fully-connected layers in our work) for the seizure prediction task.



\section{Conclusion}

Seizure prediction capability has been studied and improved over the last four decades. A perfect prediction is yet available but with current prediction performance, it is useful to provide the patients with warning message so they can take some precautions for their safety. We have shown that feature extraction for seizure prediction can be done using unsupervised deep learning or GAN particularly. Seizure prediction can be implemented efficiently on a low-power hardware. Though our working prototype that uses off-the-shelf devices does not provide impressive power consumption, it is obviously that power consumption can be greatly reduced with customized devices. This will help patients with epilepsy to have a more manageable life with a seizure prediction device.


%



\section{Acknowledgment}
O. Kavehei acknowledges support provided via a 2018 Early Career Research grant from the Faculty of Engineering and Information Technology, The University of Sydney. This research was enabled by Sydney Informatics Hub, funded by the University of Sydney.
\ifCLASSOPTIONcaptionsoff
  \newpage
\fi



%
\bibliographystyle{IEEEtran}
\bibliography{IEEEabrv,MyCollection}{}

%








\end{document}